\documentclass[11pt,a4paper]{paperstyle}

\usepackage[authoryear,compress,round]{natbib}
\usepackage{amsmath,amssymb}
\usepackage{graphicx}
\usepackage{booktabs}
\usepackage{multirow}
\usepackage{url}
\usepackage[bottom]{footmisc}
\usepackage{graphicx}
\usepackage{float}
\renewcommand{\today}{}

\runningtitle{Your Paper Title}
\institution{Your Institution}
\title{\vspace*{-0.6cm}\centering GPT-6-Astra Understands Touch and Outperform Vision--Tactile--Language--Action Models}
\institution{Sun Yat-sen University, X-Era AI Lab, The Hong Kong Polytechnic University}

\author{\vspace{-0.8cm}
Ziyong Wang,
Yupei Lin,
Runxuan Huang,
Liang Lin,
Dongyu Zhang\textsuperscript{$\dagger$},
Guangrun Wang\textsuperscript{$\dagger$}\\
\small Sun Yat-sen University, X-Era AI Lab, The Hong Kong Polytechnic University\\
\small \texttt{wangzy369@mail2.sysu.edu.cn, donny.zhang@cpce-polyu.edu.hk, wanggrun@gmail.com}\\
\small\textcolor{red}{https://gpt-as-tactile-policy.github.io/}
\vspace{-0.7cm}}

\fancypagestyle{titlepagefooter}[firststyle]{
  \fancyfoot[L]{\normalfont\footnotesize\textsuperscript{$\dagger$}Corresponding authors.}
}

\begin{abstract}
\vspace{-0.6cm}
General-purpose multimodal models have shown promise as direct robot action policies, yet whether they grasp the physics underlying real-world interactions and can apply that understanding to guide action remains underexplored. We investigate this question using contact-rich manipulation as a representative testbed, examining tactile understanding as a concrete instance of this broader capability.
Contact-rich manipulation requires interpreting tactile feedback and adjusting actions after obstruction. Existing learned tactile policies acquire these capabilities through specialized pretraining and per-task optimization, with substantial data and training costs. We present GPT-as-Tactile-Policy, which summarizes expert demonstrations into task guidance and tactile references for contact assessment and corrective action. GPT-6-Astra combines this context with multimodal observations and interaction history to generate control commands and adjust actions through contact feedback. Without additional robotics-specific pretraining or per-task policy fine-tuning, our method achieves 93.8\% mean success across five UniVTAC tasks. Overall performance is comparable to state-of-the-art policies, with higher success on selected tasks. On Insert Hole, tactile feedback improves success from 31\% to 99\%. Execution analysis further demonstrates contact-guided probing and correction, supporting manipulation progress through explicit expert experience and online feedback.
\end{abstract}

\begin{document}
\maketitle

\thispagestyle{titlepagefooter}

\begingroup
\setlength{\intextsep}{4pt}

\begin{figure}[H]
  \centering
  \setlength{\abovecaptionskip}{0pt}
  \setlength{\belowcaptionskip}{0pt}
  \includegraphics[
    width=\linewidth,
    height=0.36\textheight,
    keepaspectratio
  ]{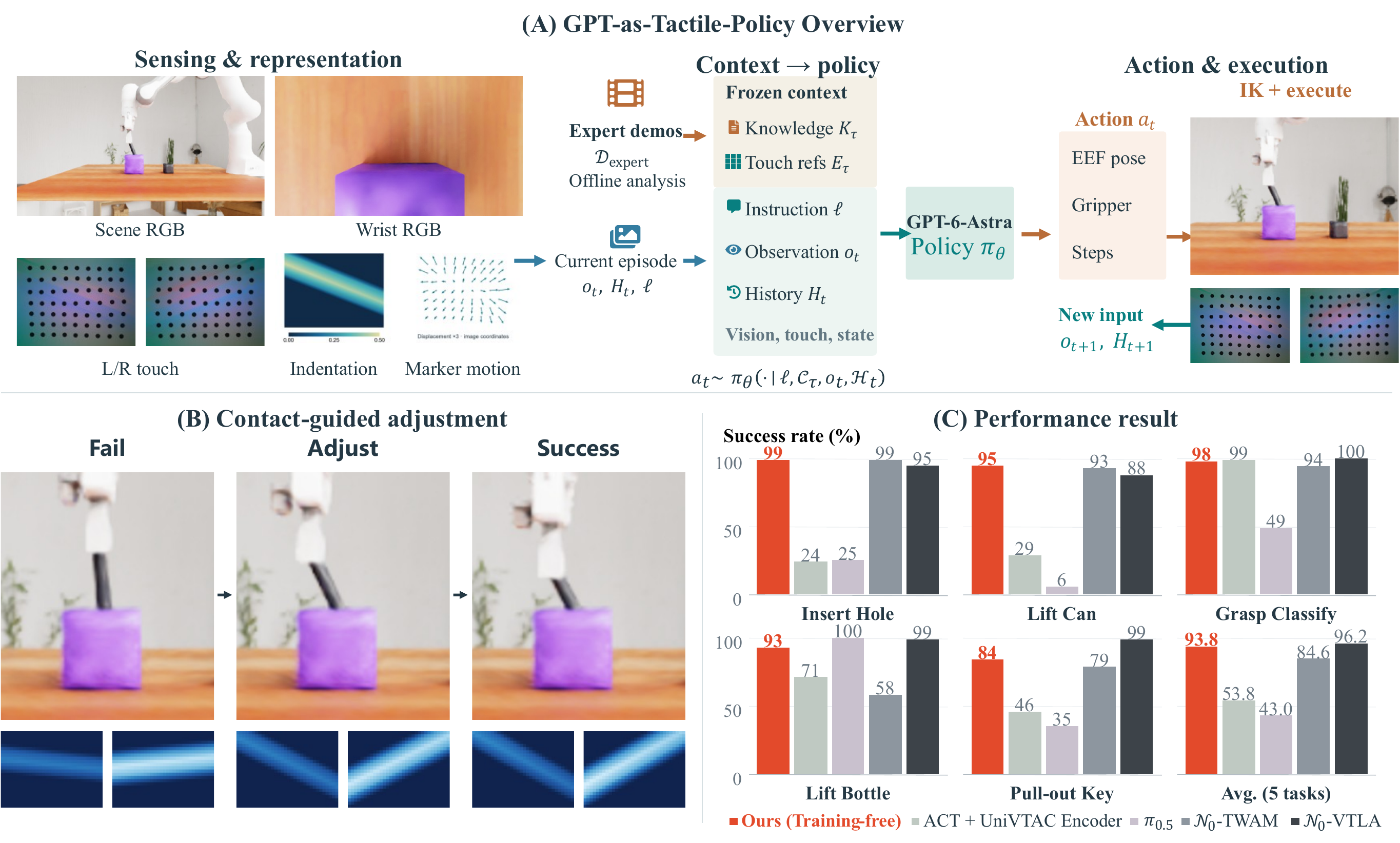}

  \caption{\textbf{GPT-as-Tactile-Policy for contact-rich manipulation.}
    \textbf{(A)} Offline task knowledge construction and online
    closed-loop control.
    \textbf{(B)} Contact-guided insertion using visual and bilateral
    tactile feedback.
    \textbf{(C)} Task-wise success rates and average performance
    on UniVTAC.}
  \label{fig:teaser}
\end{figure}

\endgroup

\section{Introduction}
\label{sec:introduction}

The role of general-purpose large models in robotics is expanding
from task planning and skill selection \citep{ahn2022saycan}
and control program generation \citep{liang2023code}
to direct action decisions. Recent evaluations of GPT-6-Astra
show that a general-purpose multimodal model can generate
numerical robot actions from observations, task instructions,
and execution history, using feedback to determine subsequent
behavior \citep{team2026systematically}. 
These studies progressively connect general knowledge to robot action,
raising a fundamental question: do general-purpose multimodal models
grasp the physics underlying real-world interactions, and can they
apply that understanding to guide their actions? Exploring this
question requires examining how models reason the physical
consequences of their actions and use interaction feedback to adjust
subsequent behavior
\citep{huang2022innermonologue,liu2023reflect}.

Tactile understanding provides a representative test case for a model's grasp of physics, because interpreting tactile feedback involves making sense of physical phenomena such as contact, deformation, and constrained motion. We therefore choose contact-rich tasks such as insertion and alignment as our experimental setting \citep{huang2025tactilevla,bi2025vlatouch}. Even after identifying a target and reaching the contact location, a robot can be obstructed by local pose errors and hidden geometric constraints \citep{huang2024_3dvitac,xue2025reactivediffusion}. External vision provides scene layout and visible object motion, but offers limited access to deformation and motion constraints at the contact interface \citep{chen2026univtac,higuera2026visuotactileworldmodels}. In this setting, tactile feedback allows us to examine whether models can connect the observed physical effects to their own actions and use that understanding to guide subsequent control.

\Needspace{3\baselineskip}
Existing research has developed learned policies that integrate
touch into contact-rich manipulation. Representative approaches
include ACT with the UniVTAC encoder \citep{chen2026univtac},
the tactile-predictive $\mathcal{N}_{0}$-VTLA
\citep{n0vtla2026}, and the tactile-native world-action model
$\mathcal{N}_{0}$-TWAM \citep{n0twam2026}. These approaches
acquire contact representations and manipulation capabilities
through representation learning, pretraining, and downstream
policy optimization. Their use of expert experience therefore
involves training dedicated representations or updating policy
parameters for downstream tasks. This motivates exploring an
alternative: making expert manipulation experience directly
available to a general-purpose model as task guidance,
without an additional task-specific policy training stage.

\begin{figure}[ht]
  \centering
  \includegraphics[width=\linewidth]{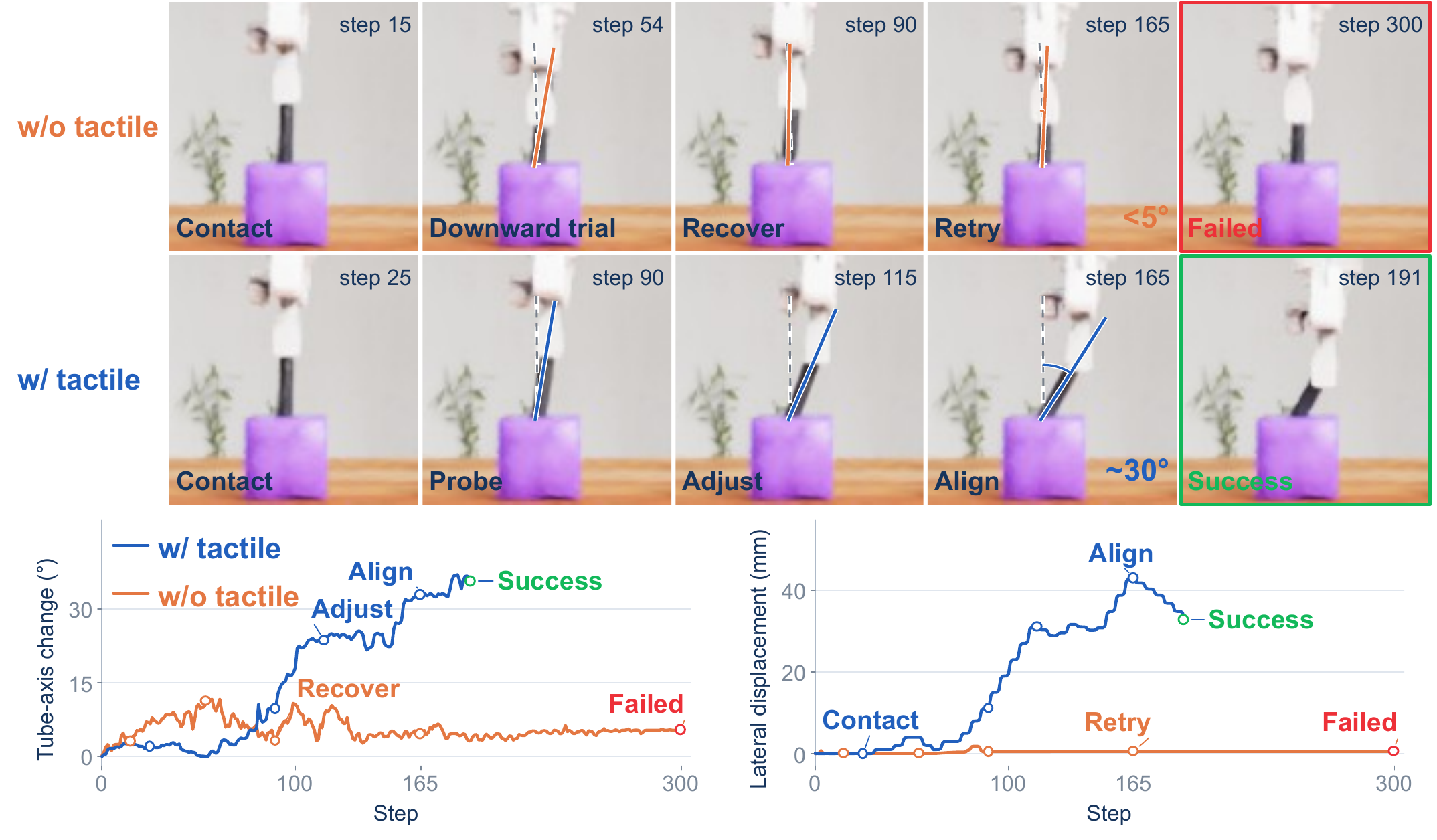}
  \caption{\textbf{Our GPT-as-Policy on Insert Hole.}
    \emph{w/o tactile} and \emph{w/ tactile} denote our policy
without and with tactile feedback, respectively.
    Keyframes and pose-change curves contrast failed retries
    with successful alignment, followed by completion of the insertion.}
  \label{fig:tube_touch_comparison}
\end{figure}

\begin{table}[h]
  \centering
  \caption{\textbf{Training costs and knowledge-guided probing.}
    Our method uses explicit expert knowledge without additional
    pretraining or per-task policy optimization.}
  \label{tab:method_design_comparison}
  \begingroup
  \fontsize{9}{12}\selectfont
  \setlength{\tabcolsep}{4pt}
  \renewcommand{\arraystretch}{1.4}
  \begin{tabular*}{\linewidth}{@{\extracolsep{\fill}}lccc@{}}
    \toprule
    \textbf{Method} & \textbf{Specialized pretraining}
      & \textbf{Per-task training} & \textbf{Knowledge-guided probing} \\
    \midrule
    ACT + UniVTAC
      & Tactile encoder & 50 demos; 4,000 steps & $\times$ \\
    $\mathcal{N}_{0}$-VTLA
      & Multi-node cluster & One policy per task & $\times$ \\
    $\mathcal{N}_{0}$-TWAM
      & $128\!\times\!$H800; $8\!\times\!$A100
      & One policy/task; 1k steps & $\times$ \\
    \midrule
    \textbf{Ours} & \textbf{No additional pretraining}
      & \textbf{No parameter updates} & $\boldsymbol{\checkmark}$ \\
    \bottomrule
  \end{tabular*}
  \endgroup
\end{table}

To investigate this question without additional specialized pretraining
or per-task policy fine-tuning, we introduce GPT-as-Tactile-Policy,
which summarizes expert manipulation experience into task guidance
that a general-purpose model can use directly. As shown in Fig.~\ref{fig:teaser}A, the model
combines this experience with current visual and tactile observations
and interaction history to directly generate robot actions, revising
subsequent attempts based on contact changes. Expert experience provides
principles for probing and correction, while real-time tactile feedback
helps the model determine how to apply them to the current interaction.
This process requires no policy parameter updates;
Tab.~\ref{tab:method_design_comparison} compares its training
requirements with those of existing methods.

Fig.~\ref{fig:teaser}B and Fig.~\ref{fig:tube_touch_comparison} illustrate how tactile feedback changes the policy's response to obstruction. Without tactile feedback, the policy repeatedly presses downward, retracts, and retries, but its orientation and lateral adjustments remain limited, leaving insertion incomplete. With tactile feedback, the policy actively explores alternative insertion directions and progressively adjusts the tube orientation and lateral position until insertion succeeds. This contrast illustrates how tactile feedback helps identify ineffective attempts, guide local correction, and restore manipulation progress. Across five contact-rich UniVTAC tasks, our method achieves overall performance comparable to extensively pretrained state-of-the-art methods and higher success rates on selected tasks, without additional robotics-specific pretraining or per-task policy fine-tuning, as shown in Fig.~\ref{fig:teaser}C. Our main contributions are:

\begin{itemize}
  \item \textbf{Direct visuo-tactile control.}
  We introduce GPT-as-Tactile-Policy, enabling a fixed-parameter
  general-purpose multimodal model to translate tactile feedback into
  effective action adjustments. By relating contact changes to preceding
  motions, the model revises its approach and pose to improve alignment
  and make progress after obstruction.

  \item \textbf{Experience-guided tactile reasoning.}
  We summarize expert manipulation experience into fixed task guidance
  covering contact cues, probing, and corrective strategies. Together
  with representative tactile references, this guidance helps the model
  apply prior experience to current interactions.

  \Needspace{8\baselineskip}
  \item \textbf{SOTA-level performance without policy fine-tuning.}
  Across five UniVTAC tasks, evaluated over 100 seeds per task
  following the benchmark protocol, our method achieves overall
  performance comparable to extensively pretrained state-of-the-art
  methods and higher success rates on selected tasks, without additional
  robotics-specific pretraining or per-task policy fine-tuning.
  Execution analyses further demonstrate how the model uses tactile
  feedback to select probing directions, correct its pose, and make
  progress after obstruction.
\end{itemize}

\section{Related Work}
\label{sec:related}

\subsection{Language Models as Robot Policies}
\label{sec:related_lm_policies}

Language models have been used to translate pretrained task knowledge
into robot behavior through skill selection, program generation, and
motion planning. SayCan~\cite{ahn2022saycan} combines language-model
estimates of task relevance with value-function estimates of skill
feasibility to select skill sequences for long-horizon manipulation.
Code as Policies~\cite{liang2023code} generates policy
programs that process perception outputs and parameterize control
primitives, using program logic to express motion parameters,
conditional behavior, and feedback loops.
Instruct2Act~\cite{huang2023instruct2act} maps multimodal instructions
to Python policy programs that invoke perception, planning, and
action interfaces.
RoboScript~\cite{chen2024roboscript} emphasizes the deployability of
generated programs by integrating perception, motion planning, and
control through a unified interface for simulation and real robots.
VoxPoser~\cite{huang2023voxposer} grounds language-inferred affordances
and constraints in visual observations through composable 3D value
maps, which support closed-loop trajectory generation through
model-based planning.
These approaches connect language-model knowledge to robot execution
through different intermediate representations.

Execution feedback further enables language models to revise their
decisions based on observed outcomes.
Inner Monologue~\cite{huang2022innermonologue} incorporates success
detection, scene descriptions, and human interaction as language
feedback to update task plans during execution.
REFLECT~\cite{liu2023reflect} constructs hierarchical summaries of
robot experiences from multisensory observations, uses a language
model to explain failures, and provides these explanations to a
language-based planner for corrective planning.
These studies incorporate the consequences of previous actions
into subsequent decisions.

Another line of work investigates language models that generate
explicit action parameters.
Natural Language as Policies~\cite{mikami2024natural} reasons
over textual descriptions of tasks and scene objects to generate
coordinate-level parameters for pick-and-place or sweeping actions,
reducing reliance on intermediate policy code.
GPT-as-Policy~\cite{team2026systematically} systematically evaluates
GPT-6-Astra as an embodied policy across multiple domains, examining
numerical action generation, cooperation with learned policies,
and feedback-driven adaptation.
Following this direct action-decision paradigm, We study this direct-policy paradigm in contact-rich manipulation,
focusing on how a general-purpose multimodal model uses
visual--tactile feedback to interpret contact and select subsequent
actions.

\subsection{Visuo-Tactile Manipulation Policies}
\label{sec:related_visuotactile}

Visuo-tactile manipulation combines scene observations with local
contact feedback to support fine-grained perception and control.
VITaL Pretraining~\cite{george2025vital} investigates visuo-tactile
pretraining for imitation learning, showing benefits for both
policies that use touch during execution and policies deployed
without tactile input.
3D-ViTac~\cite{huang2024_3dvitac} integrates visual and tactile
observations into a unified 3D representation that preserves their
spatial structure and relationships, and combines this representation
with diffusion policies for fine-grained manipulation.
Reactive Diffusion Policy~\cite{xue2025reactivediffusion} introduces
a slow-fast hierarchy that combines latent action-chunk generation
with high-frequency tactile feedback control, allowing the policy
to respond to contact changes during chunk execution.
These studies investigate how tactile information supports skill
learning and execution through representation learning and
feedback-aware control architectures.

Recent work further integrates tactile sensing with the task
knowledge of vision-language-action (VLA) models.
VLA-Touch~\cite{bi2025vlatouch} keeps the base VLA fixed while using
a pretrained tactile-language model to provide semantic feedback
for high-level planning and a diffusion-based controller to refine
actions using tactile signals.
Tactile-VLA~\cite{huang2025tactilevla} combines tactile sensing with
hybrid position-force control to connect vision-language knowledge
to physical interaction; its CoT variant further explores
tactile-feedback-driven failure analysis and strategy adaptation.
OmniVTLA~\cite{cheng2026omnivtla} addresses heterogeneous tactile
inputs through a dual-path tactile encoder and semantically
aligned representations.
$N_0$-VTLA~\cite{n0vtla2026} further explores large-scale
visuo-tactile pretraining, a predictive latent tactile pathway,
and offline policy improvement from stored deployment data.
Together, these approaches incorporate touch into semantic
understanding, policy adaptation, and action generation at
different decision levels.

Another line of work uses predictions of physical interactions
to support manipulation.
Visuo-Tactile World Models~\cite{higuera2026visuotactileworldmodels}
incorporate touch into action-conditioned world models to improve
the physical consistency of imagined futures and support robot
planning.
Tactile-WAM~\cite{wu2026tactilewam} introduces asymmetric attention
that isolates the visual prediction pathway from tactile tokens
while preserving tactile access for action generation.
VT-WAM~\cite{tian2026vtwam} jointly learns future visual,
tactile-deformation, and action prediction, and uses training-time
contact-gated attention guidance to strengthen the use of tactile
evidence in action generation.
$N_0$-TWAM~\cite{n0twam2026} further combines large-scale
visuo-tactile joint training, a unified force-based tactile
representation, and contact-event-based task staging to support
multi-stage manipulation.
These studies extend the role of touch from observing the current
contact state to anticipating how contact evolves under robot
actions.

Our work studies a complementary setting in which a
general-purpose multimodal model serves as a direct action
policy with its parameters fixed.
Task knowledge is supplied through in-context guidance, and
the model combines this guidance with visual--tactile
observations and execution history to select subsequent actions.
Our focus is on how the model interprets contact feedback and
uses it to make online action decisions, complementing research
on learned visuo-tactile representations, control architectures,
and predictive models.

\section{Method}
\label{sec:method}

\subsection{GPT-as-Tactile-Policy}
\label{sec:overview}

GPT-as-Policy uses a general-purpose multimodal model as a closed-loop robot action policy, selecting actions from the task instruction, observations, and interaction history \citep{team2026systematically}. In its direct-control formulation, GPT generates numerical end-effector and gripper targets, which a generic inverse-kinematics controller converts into robot motion; new observations after execution return to the model. The model therefore selects the next operation using available observations and feedback from preceding actions.

For contact-rich manipulation, we incorporate bilateral tactile feedback and task guidance summarized from expert demonstrations into this decision process. As shown in Fig.~\ref{fig:teaser}A, the framework comprises offline experience summarization and online feedback control. Offline, GPT-6-Astra jointly analyzes expert actions, contact changes, and task progress to construct fixed context $\mathcal C_\tau$ for task $\tau$. Online, the same model serves as the sole action policy, combining task guidance, visuo-tactile observations, and robot proprioception to interpret current contact responses and select actions. The resulting policy uses both fixed task experience and observations and history that evolve throughout the current interaction.

We formulate this policy as conditional action generation. At step $t$, the observation is denoted by $o_t=(V_t,s_t,\mathcal T_t)$, where $V_t$ contains scene and wrist-camera images, $s_t$ denotes the robot's proprioceptive state, and $\mathcal T_t$ comprises tactile observations from fingertips. Given a natural-language task instruction $\ell$, the policy generates an action as
\begin{equation}
  a_t\sim\pi_\theta\!\left(\cdot\mid\ell,\mathcal C_\tau,o_t,\mathcal H_t\right),
  \label{eq:policy}
\end{equation}
where $\mathcal C_\tau$ denotes the fixed task context comprising a textual Task Skill and reference tactile observations, $\mathcal H_t$ records preceding observations, actions, and execution feedback within the episode, and $a_t$ specifies the target end-effector pose, gripper position, and a short execution horizon. After each execution, new observations enter the interaction history and provide feedback for subsequent actions. Model parameters $\theta$ remain fixed throughout offline experience summarization and online execution.

\subsection{Task Knowledge Construction from Expert Demonstrations}
\label{sec:knowledge}

To provide task guidance for the online policy in Section~\ref{sec:overview}, we construct fixed task context from expert demonstrations before execution. These demonstrations contain sequential visual and tactile observations, robot states, and expert actions, recording changes in contact and task progress during manipulation. As shown in Fig.~\ref{fig:teaser}A, GPT-6-Astra jointly analyzes these sequences, relating contact changes and object motion before and after actions to summarize how experts assess progress, select adjustments, and evaluate their effects. The resulting guidance forms the textual Task Skill $K_\tau$.

The Task Skill organizes this experience into task objectives, manipulation principles, evidence for stage transitions, and corrective strategies for obstructions. It specifies which observations to examine and how they should inform subsequent actions. For example, increasing deformation without object advancement during insertion prompts retraction to relieve contact, followed by a positional or angular adjustment evaluated through subsequent feedback. This knowledge gives tactile changes operational meaning, helping the model determine when to advance, when to adjust, and which feedback to examine when assessing an adjustment.

GPT-6-Astra also extracts representative tactile observations from expert demonstrations to form the reference set $\mathcal E_\tau^{\mathrm{tac}}$, preserving bilateral contact distributions and local deformation patterns. The Task Skill provides principles for contact assessment and manipulation adjustment, while tactile references provide perceptual grounding for these principles. We organize both components into the task context
\begin{equation}
  \mathcal C_\tau=\left(K_\tau,\mathcal E_\tau^{\mathrm{tac}}\right),
  \label{eq:task_context}
\end{equation}
which serves as a fixed conditioning input to the GPT policy defined in Section~\ref{sec:overview}. At each decision step, the model combines this context with current observations and interaction history to assess whether tactile changes accompany manipulation progress and to select continued advancement, probing, or corrective actions. New observations after execution reveal the effects of the selected action and inform the next decision.

\subsection{Online Visuo-Tactile Observation Representation}
\label{sec:observations}

During online execution, we provide both bilateral tactile images and structured tactile descriptions to preserve local contact details and support comparison of contact distributions and deformation changes. Following Section~\ref{sec:overview}, we expand $\mathcal T_t$ as
\begin{equation}
  \mathcal T_t=\left\{\left(I_t^{f},D_t^{f},M_t^{f}\right)\right\}_{f\in\{L,R\}},
  \label{eq:tactile_representation}
\end{equation}
where $f\in\{L,R\}$ indexes the left and right fingertips, respectively, and $I_t^{f}$ denotes the tactile image. The descriptors $D_t^{f}$ summarize the spatial distribution and statistics of indentation depth, while $M_t^{f}$ records two-dimensional marker displacements relative to reference positions.

For spatial contact distributions, we spatially pool gel indentation maps and provide statistical descriptions, including mean indentation, peak indentation, contact regions, and centroids, characterizing contact location, extent, and indentation distribution. Marker displacements further describe local deformation directions and spatial patterns within contact regions. Information from the two fingertips is retained separately, allowing the model to compare bilateral contact differences and their evolution.

To relate contact changes to executed actions, the model considers indentation distributions and marker displacements across successive observations together with the measured end-effector and gripper motion over the same interval. Visual observations further reveal object motion relative to the target and gripper, connecting actual robot motion, object response, and tactile changes. Updated through execution, these observations provide current contact evidence for applying the Task Skill and feedback for subsequent closed-loop action generation.

\subsection{Closed-Loop Action Generation from Interaction Feedback}
\label{sec:control}

During online execution, the decision policy $\pi_\theta$ assesses manipulation progress using the current observation $o_t$ and interaction history $\mathcal H_t$. It constructs action $a_t$ from the latest measured robot state. After checks on command bounds and format, a generic inverse-kinematics controller executes the target for one to five control steps. A new observation $o_{t+1}$ is then acquired. Short execution horizons allow prompt contact assessment and adjustment of subsequent motion.

After each execution, the decision policy $\pi_\theta$ compares $o_t$ and $o_{t+1}$ to assess the effects of action $a_t$. It jointly examines end-effector motion, object responses, and tactile changes to determine whether manipulation has progressed. Guided by Task Skill $K_\tau$, the policy determines whether alignment or grasp adjustments are needed. It then generates the next action $a_{t+1}$.

When observations leave the adjustment direction uncertain, the decision policy $\pi_\theta$ actively explores to acquire additional contact information. Exploratory actions include small translations, orientation adjustments, and changes in gripper opening. The policy assesses contact changes and task progress using object motion and bilateral tactile feedback. This feedback guides the direction and magnitude of subsequent actions. Exploratory actions and their observed outcomes are incorporated into the interaction history $\mathcal H_t$. Subsequent decisions therefore build on accumulated interaction evidence to progressively refine manipulation.

\section{Experiments}
\label{sec:experiments}

We compare our method with visual, visuo-tactile, and world-action
baselines on five UniVTAC tasks and conduct a tactile ablation
on Insert Hole.

\subsection{Experimental Setup}
\label{sec:experimental_setup}

\paragraph{Tasks and data.}
We evaluate on five contact-rich manipulation tasks from UniVTAC \citep{chen2026univtac}: \emph{Insert Hole}, \emph{Lift Can}, \emph{Grasp Classify}, \emph{Lift Bottle}, and \emph{Pull-out Key}. These tasks cover geometric alignment, grasped-object manipulation, constrained motion, and tactile classification.

For each task, we use 50 expert demonstrations, split at the episode
level into 40 for knowledge construction and 10 for offline validation.
From the construction subset, GPT-6-Astra analyzes contact changes and
object motion before and after expert actions to distill the Task Skill
for progress assessment and manipulation adjustment. It also extracts tactile references
preserving bilateral contact distributions and local deformation
patterns. Together, these components form the task context, which is
fixed before formal evaluation and used across all test episodes
of the same task.

\paragraph{Baselines.}
We compare against four methods previously evaluated on UniVTAC. E0 \citep{zhan2026e0}and $\pi_{0.5}$ provide a vision-only baseline to assess manipulation under contact constraints \citep{intelligence2025pi_}. ACT + UniVTAC Encoder uses pretrained visuo-tactile representations \citep{chen2026univtac}. $\mathcal N_0$-VTLA is a state-of-the-art policy with latent tactile prediction \citep{n0vtla2026}. $\mathcal N_0$-TWAM jointly models future vision, touch, and actions \citep{n0twam2026}.

\paragraph{Policy configuration.}
We use GPT-6-Astra as the decision policy $\pi_\theta$, with reasoning effort fixed to \texttt{xhigh}. The same model and reasoning configuration is used across all tasks. Each test episode starts an independent model session, with interaction history $\mathcal H_t$ updated within the episode and reset between episodes. Observations and action execution follow Sections~\ref{sec:observations} and~\ref{sec:control}, respectively.

\paragraph{Evaluation protocol.}
Our evaluation comprises 100 fixed test episodes per task, for a total of
500 episodes. Episode outcomes are determined by the environment's native
success criteria. We report per-task success rates and the average
over the five tasks, assigning equal weight to each task. Terminated
episodes that do not satisfy the success conditions are counted as
failures.

\subsection{Main Result}
\label{sec:main_comparison}

Tab.~\ref{tab:univtac_main} compares GPT-as-Policy with ACT + UniVTAC Encoder, $\pi_{0.5}$, $\mathcal N_0$-VTLA, and $\mathcal N_0$-TWAM on five UniVTAC tasks. Our GPT-as-Policy achieves competitive performance without additional robotics-specific pretraining or per-task policy fine-tuning. Its mean success rate of 93.8\% approaches the 96.2\% achieved by $\mathcal N_0$-VTLA and exceeds the 84.6\% of $\mathcal N_0$-TWAM. Viewed alongside the training requirements in Tab.~\ref{tab:method_design_comparison}, these results highlight a different route to task adaptation: expert demonstrations are organized into task guidance that a general-purpose model can apply during online control.

On Insert Hole, GPT-as-Policy achieves 99\% success, matching $\mathcal N_0$-TWAM and outperforming $\mathcal N_0$-VTLA. On Lift Can, it achieves 95\%, the highest success rate among the compared methods. Across all five tasks, it matches or outperforms $\mathcal N_0$-TWAM, demonstrating consistent performance across tasks.

\begin{table}[t]
  \centering
  \caption{Success rates on five UniVTAC tasks, in percent.
  Mean denotes the task average.}
  \label{tab:univtac_main}
  \smallskip
  \small
  \setlength{\tabcolsep}{4pt}
  \renewcommand{\arraystretch}{1.15}
  \begin{tabular*}{\linewidth}{@{\extracolsep{\fill}}lcccccc@{}}
    \toprule
    Method & Insert Hole & Lift Can & Grasp Classify
    & Lift Bottle & Pull-out Key & Mean \\
    \midrule
    ACT + UniVTAC Encoder & 24.0 & 29.0 & 99.0 & 71.0 & 46.0 & 53.8 \\
    $\pi_{0.5}$ & 25.0 & 6.0 & 49.0 & 100.0 & 35.0 & 43.0 \\
    $\mathcal{N}_{0}$-VTLA & 95.0 & 88.0 & 100.0 & 99.0 & 99.0 & 96.2 \\
    $\mathcal{N}_{0}$-TWAM & 99.0 & 93.0 & 94.0 & 58.0 & 79.0 & 84.6 \\
    \midrule
    \textbf{Ours} & \textbf{99.0} & \textbf{95.0} & \textbf{98.0}
    & \textbf{93.0} & \textbf{84.0} & \textbf{93.8} \\
    \bottomrule
  \end{tabular*}
\end{table}

\begin{figure}[!htb]
  \centering
  \includegraphics[width=\linewidth]{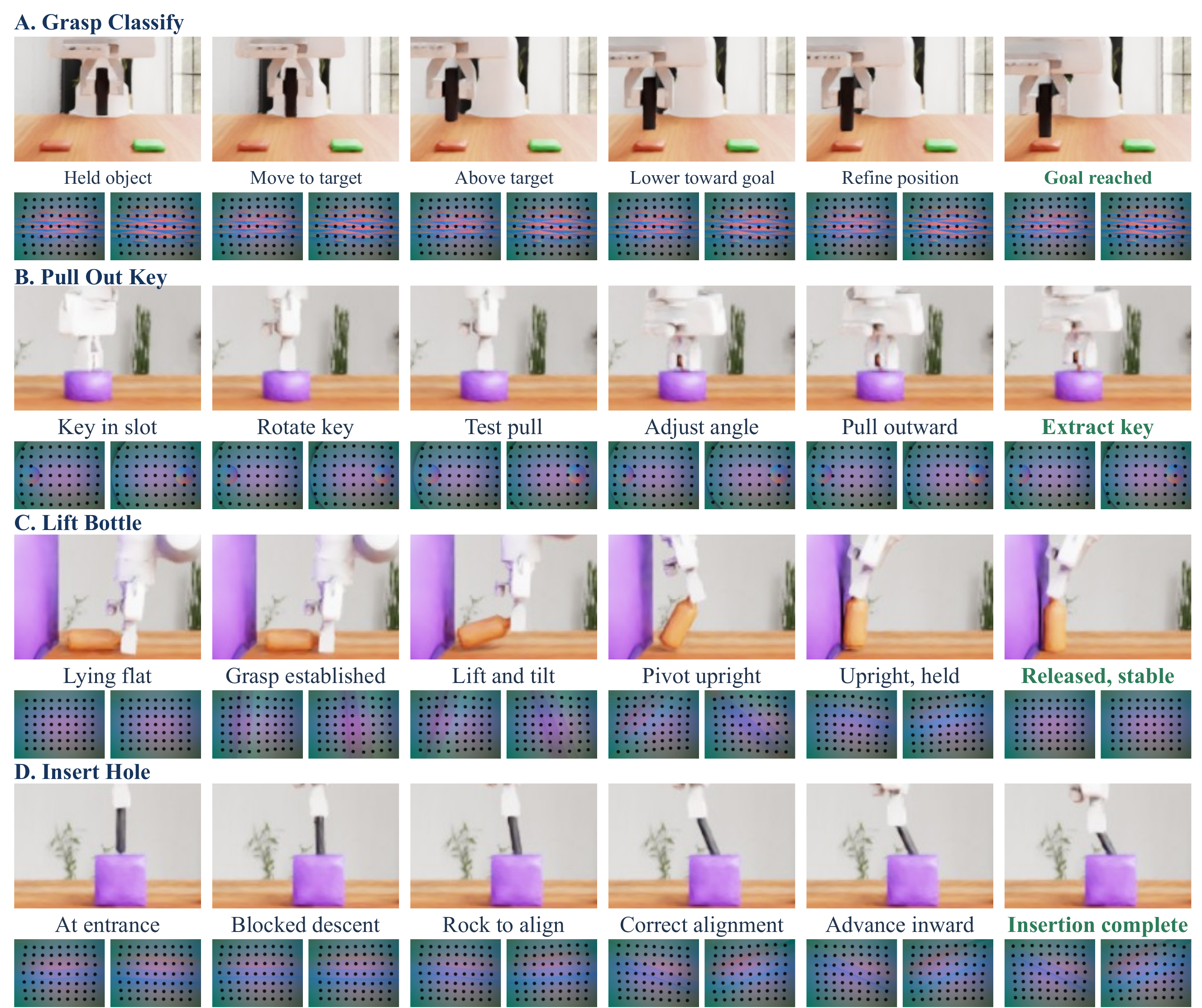}
  \caption{\textbf{Successful executions across four tasks.}
    \textbf{(A)} Grasp Classify. \textbf{(B)} Pull-out Key.
    \textbf{(C)} Lift Bottle. \textbf{(D)} Insert Hole.
    Selected keyframes at representative timesteps are arranged
    chronologically from left to right. Beneath each RGB frame, the left
    and right images show the corresponding tactile observations from
    the left and right fingertips, respectively.}
  \label{fig:successful_executions}
\end{figure}

\paragraph{Qualitative Results.}
Fig.~\ref{fig:successful_executions} presents successful executions on Grasp Classify, Pull-out Key, Lift Bottle, and Insert Hole, together with bilateral tactile observations. In Grasp Classify, the policy refines the object position near the target region to complete placement. In Pull-out Key, a test pull is followed by an angle adjustment, outward motion, and extraction. In Lift Bottle, the policy establishes a grasp, lifts and reorients the bottle, and releases it in a stable pose. In Insert Hole, the policy adjusts the tube alignment after a blocked descent, then advances inward to complete insertion. These sequences illustrate how the decision policy $\pi_\theta$ completes manipulation through successive adjustments under different contact constraints.

\subsection{Ablation on Tactile Information}
\label{sec:tactile_ablation}

We evaluate the role of tactile feedback in corrective actions and reliable insertion on Insert Hole. We compare the full policy \emph{Ours} with a tactile-free variant \emph{w/o tactile}. The variant retains vision, proprioception, and the action interface, while removing tactile information and tactile references. Each configuration uses a fixed Task Skill matched to its available observations. Both are evaluated on the same 100 test configurations with identical initialization, low-level controllers, and execution budgets. Model and session settings follow Section~\ref{sec:experimental_setup}.

Tab.~\ref{tab:tactile_ablation} shows that, on Insert Hole, the full policy achieves 99\% success, compared with 31\% for \emph{w/o tactile}. This represents an improvement of 68 percentage points. The gap highlights the difficulty of reliable insertion using vision and proprioception alone. Tactile feedback substantially improves task completion under the evaluated contact constraints.

\begin{table}[t]
  \centering
  \caption{Tactile ablation on Insert Hole. V, P, and T denote vision, proprioception, and touch.}
  \label{tab:tactile_ablation}
  \smallskip
  \small
  \setlength{\tabcolsep}{6pt}
  \renewcommand{\arraystretch}{1.15}
  \begin{tabular*}{\linewidth}{@{\extracolsep{\fill}}lcc@{}}
    \toprule
    Method
    & Online observations
    & Success rate (\%) $\uparrow$ \\
    \midrule
    Ours        & V + P + T & \textbf{99.0} \\
    w/o tactile & V + P     & 31.0 \\
    \bottomrule
  \end{tabular*}
\end{table}

\paragraph{Execution behavior with and without tactile feedback.}
To illustrate how tactile feedback affects adjustments after obstruction, we visualize representative insertion episodes with and without touch in Fig.~\ref{fig:tube_touch_comparison}. Without touch, repeated attempts produce limited changes in tube orientation and lateral end-effector position. Insertion remains incomplete at step~300. With touch, the policy actively explores alternative insertion directions and progressively adjusts the tube orientation and lateral position. It completes insertion at step~191. The angle and displacement traces record these adjustments relative to each episode's initial state. The tactile-enabled policy revises its local approach, illustrating how contact feedback supports exploration of alternative insertion directions and restores manipulation progress.

\subsection{Qualitative Analysis on Insert Hole}
\label{sec:qualitative_insert_hole}

\begin{figure}[!htb]
  \centering
  \includegraphics[width=\linewidth]{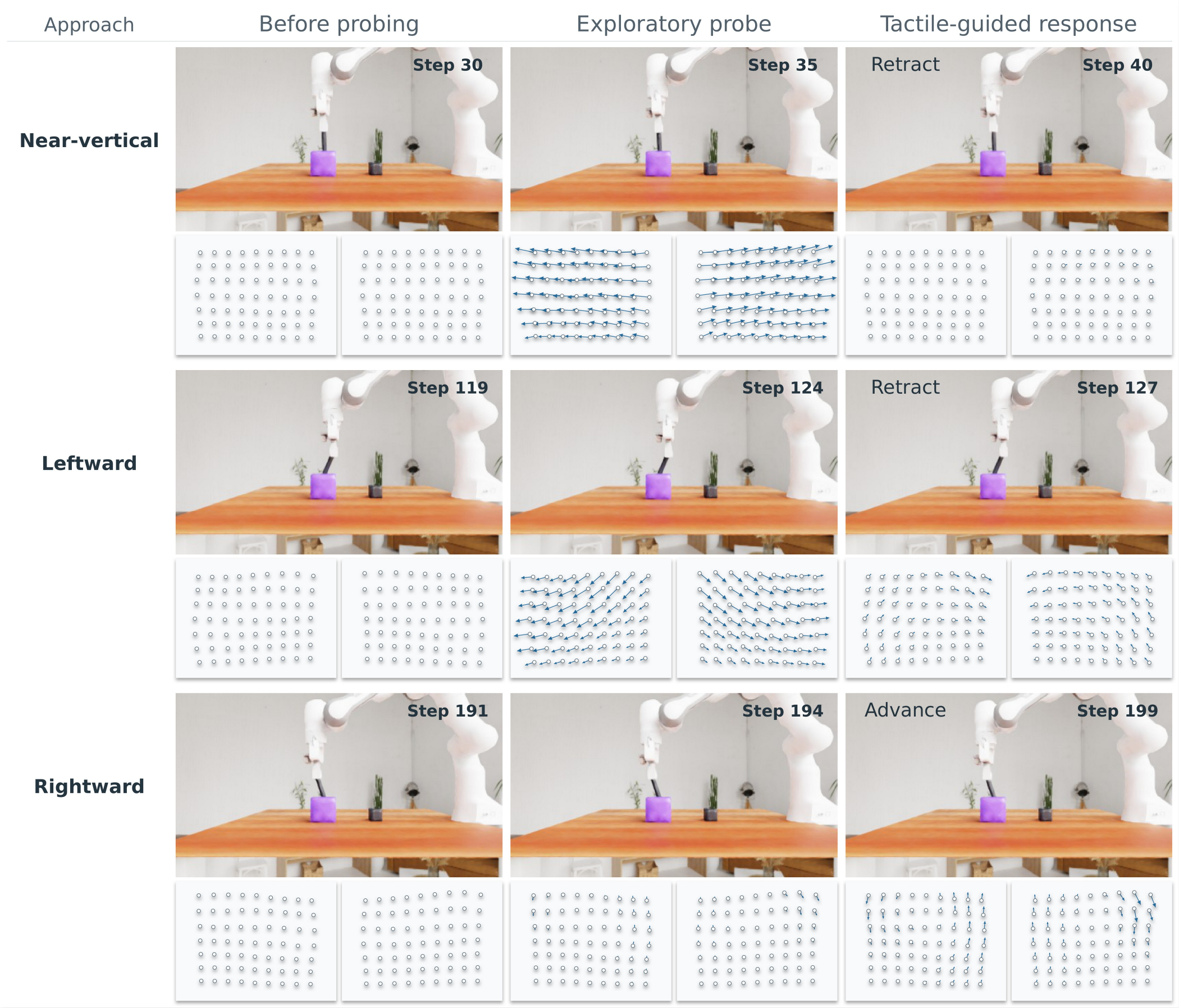}
  \caption{\textbf{Tactile-guided exploration on Insert Hole.}
    Scene images and bilateral marker-displacement fields show exploratory
    probing followed by retraction for near-vertical and leftward approaches,
    or advancement for the rightward approach.}
  \label{fig:insert_hole_exploration}
\end{figure}

To examine how tactile feedback guides active exploration, we visualize three probing sequences from a single Insert Hole episode in Fig.~\ref{fig:insert_hole_exploration}. They correspond to near-vertical, leftward, and rightward approaches. Each sequence shows the state before probing, the bilateral tactile response, and the subsequent action.

For the near-vertical approach, steps 30, 35, and 40 show the states and corresponding contact responses before probing, during probing, and after retraction. The policy then retracts, and displacement decreases. The leftward attempt at steps 119, 124, and 127 follows a similar pattern. In both attempts, withdrawal accompanies reduced marker displacement, showing that motion adjustment can relieve the local deformation induced by probing. These responses provide evidence for ending the current attempt and reconsidering the approach direction.

The rightward attempt at steps 191, 194, and 199 exhibits a different response. Probing produces smaller marker displacement, followed by continued advancement. Here, tactile feedback and subsequent motion jointly support direction selection: smaller contact changes accompany further advancement, supporting retention of the current approach. Comparing actions and outcomes across directions progressively identifies a local motion that permits insertion under the observed contact conditions.

These sequences illustrate the information-gathering role of active exploration in contact-rich manipulation. Probing reveals contact responses to candidate directions, while withdrawal enables realignment. The decision policy $\pi_\theta$ uses this feedback to withdraw or advance, restoring progress after obstruction. The responses retained in $\mathcal H_t$ also inform subsequent local action adjustments within the same episode.

\section{Conclusion}
To enable general-purpose multimodal models to use contact feedback for action adjustment, we present GPT-as-Tactile-Policy, 
which combines expert-derived task guidance with online tactile observations. Expert experience provides task guidance for contact assessment and manipulation adjustment. 
Online tactile feedback provides evidence for evaluating action outcomes and selecting subsequent adjustments. 
Without additional robotics-specific pretraining or per-task policy fine-tuning, our method achieves overall performance comparable to state-of-the-art policies across five UniVTAC tasks, matching or exceeding the compared methods on selected tasks. 
Tactile ablation and execution analysis further demonstrate the role of contact feedback in local probing, directional adjustment, and restoring progress after obstruction. These results establish explicit task knowledge combined with online interaction feedback as an effective route to efficient contact-rich manipulation.

\bibliography{references}

\end{document}